# Multiple Instance Learning with random sampling for Whole Slide Image Classification

H. Keshvarikhojasteh*[a], J.P.W. Pluim[a], M. Veta[a]
[a]Dept. of Biomedical Engineering, Eindhoven University of Technology, Eindhoven, The Netherlands

**ABSTRACT**

In computational pathology, random sampling of patches during training of Multiple Instance Learning (MIL) methods is computationally efficient and serves as a regularization strategy. Despite its promising benefits, questions concerning performance trends for varying sample sizes and its influence on model interpretability remain. Addressing these, we reach an optimal performance enhancement of 1.7% using thirty percent of patches on the CAMELYON16 dataset, and 3.7% with only eight samples on the TUPAC16 dataset. We also find interpretability effects are strongly dataset-dependent, with interpretability impacted on CAMELYON16, while remaining unaffected on TUPAC16. This reinforces that both the performance and interpretability relationships with sampling are closely task-specific. End-to-end training with 1024 samples reveals improvements across both datasets compared to pre-extracted features, further highlighting the potential of this efficient approach.

## 1. INTRODUCTION

In histopathology image analysis, the sheer size of whole-slide (WSI) images presents a bottleneck for obtaining high-quality localized annotations. The large image size also presents a computational bottleneck for the end-to-end training of models due to memory constraints. Multiple instance learning (MIL) [1] enables weakly supervised training of deep neural networks by using a single label for a bag of instances, rather than separate labels for each instance such as pixel-level or patch-level annotations commonly found in imaging tasks.

While the lack of labels can be addressed using MIL in combination with more readily available slide-level labels and treating each slide as a bag of patches, computational difficulties for end-to-end training remain. As a result, most MIL methods applied to WSIs extract features from each patch using a fixed, pretrained network, such as ResNet-50 [2] trained on ImageNet. While this gives good results, it is suboptimal as the extracted features are not specifically tailored toward representations of histological structures.

Randomly sampling a small number of patches at training time enables end-to-end training of deep neural networks by curtailing the number of patches taken from each slide per training iteration, making backpropagation computationally tractable [3]. One study [4] compared various sampling strategies for WSI classification using the CLAM backbone [5] (a multi-class MIL method with an attention mechanism) with pre-extracted features in terms of computing time and performance, with sampling ranging from 10 to 1000 instances. They found that a simple random sampling technique outperforms other strategies while reducing training time. The authors further claim that random sampling acts as a regularization technique, preventing overfitting and improving performance.

In this article, we delve deeper into the impact of random sampling on the performance for WSI classification using the CLAM backbone with pre-extracted features. We expand on previous work by conducting a comprehensive analysis of the network's performance for a wide spectrum of downsampling percentages, ranging from 2% to 100%, showing that the trend of the performance is data specific. Furthermore, we investigate the effect of random sampling on the interpretability of the method using the generated attention maps. Finally, we explore the performance enhancements achieved by incorporating end-to-end training of the CLAM method alongside random sampling.

* h.keshvarikhojasteh@tue.nl

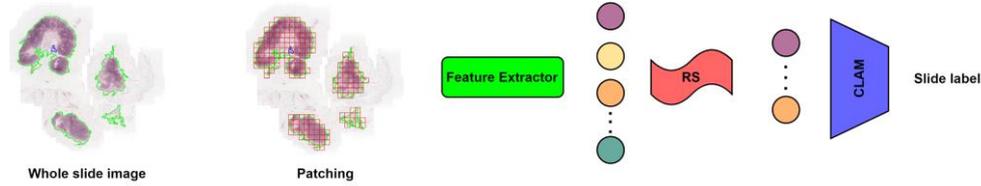

Figure 1. Overview of the proposed Framework. Firstly, the input WSI is tiled into small patches. Next, a few samples are randomly chosen, and their features are extracted using the feature-extractor. Finally, the features are fed into the CLAM to predict the slide label.

## 2. MATERIALS AND METHODS

### 2.1 Dataset

We perform all experiments with two publicly available WSI datasets, specifically CAMELYON16 [6] and TUPAC16 [7]. CAMELYON16 contains 399 H&E slides with breast cancer lymph node metastasis annotations, split into 270 slides for training and validation and 129 for testing. TUPAC16 consists of 821 H&E slides with breast tumor proliferation labels, split into 500 for training and validation, and 321 slides for testing. We dichotomize the labels of this dataset into two classes: low-grade (slides labelled as grade 1) and high-grade (slides labelled as grade 2 or 3). For both datasets, we adhere to the formal split for model development and held-out test set.

### 2.2 MIL for WSI classification

Computational pathology has widely used multiple instance learning (MIL) with various adaptations [8, 9]. In early studies, researchers proposed hard key patch selection methods such as top-$K$ [8]. However, Ilse et al. [9] developed an attention-based MIL method for binary classification that provides interpretability through attention weights. To extend the proposed method for multi-class classification, Lu et al. [5] devised the clustering-constrained attention multiple-instance learning (CLAM) method, comprising $N$ parallel attention branches for $N$ classes. The attention branches perform permutation invariant weighted averages of patch features, and their scores are computed using several fully connected layers, enabling the summarization of $N$ distinct slide representations. Furthermore, to address the issue of data inefficiency, a clustering layer is introduced on an instance-wise basis to assign high- and low-attended patches to separate clusters. Given the absence of patch-level labels, this layer is supervised by generating pseudo labels using the attention scores.

The utilization of random sampling during training of deep networks in combination with MIL has been previously explored in tasks involving both natural images [10] and medical images [11], including histopathology [3, 4]. Although seemingly straightforward, this technique allows for end-to-end training of MIL with larger bags. This is achieved by providing a smaller subset of samples to the network during each iteration, which in turn reduces the memory footprint of the training process.

### 2.3 Implementation details

We used the same experimental setup for both the TUPAC16 and CAMELYON16 datasets. We extracted non-overlapping 256×256 patches at 40× magnification. The pre-trained ResNet50 model was used to extract features, resulting in 1024-dimensional embeddings for each patch. Unless otherwise specified, we used the same hyperparameters as [5] and conducted two main experiments. Firstly, we extensively analyzed the influence of random sampling for training the CLAM network, denoted by the RS symbol in the results. Secondly, we explored the end-to-end training of CLAM with the aid of random sampling, denoted with E2E in the results. We utilized random sampling only during training time for all experiments (i.e. all patches from given slide are used at inference time).

For the RS experiment, we randomly selected patches from each slide in every iteration, based on the sampling percentage. We conducted a series of experiments, starting from the minimum number of patches required to train the model (RS-8-samples) to the all samples (RS-100%). For the E2E experiment, we trained the network in two steps to reduce training time. In the first step, we trained the CLAM model while freezing the feature-extractor model. In the second step, we unfroze the layers of the feature-extractor network from the last sub-layer of the second block onwards, except the batch-normalization layers. We then initialized the CLAM with the weights of the selected checkpoint with the lowest validation loss and trained the whole network end-to-end. We utilized various data augmentation techniques, including horizontal and vertical flips, random rotation, color jitter, and elastic deformation. We used the same set of samples as in the first step to only evaluate the end-to-end training effect. We utilized 1024 instances to conduct the experiments. In the second step, we used a ten times smaller learning rate of 2e-5 because the model was unable to converge at the initial high learning rate of 2e-4. We used k-fold cross-validation and reported the mean and standard deviation of the area under the curve (AUC) metric for each experiment, where k = 10 for RS experiments, while k = 1 for E2E experiments. The overview of the proposed method is presented in Figure 1.

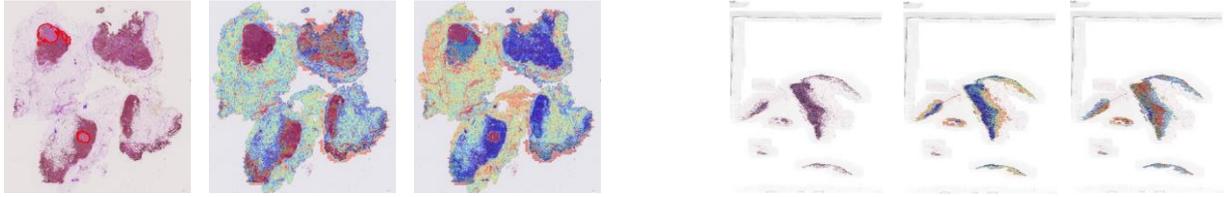

Figure 2. A tumor slide (left) and a normal slide (right) from CAMELYON16, correctly classified by both models (CLAM and RS-30%). From left to right, the slide, attention heatmap generated by the CLAM, and attention heatmap produced by the RS-30%.

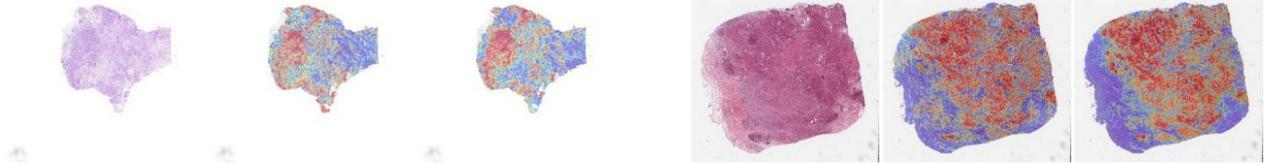

Figure 3. A negative slide (left) and a positive slide (right) from TUPAC16, correctly classified by both models (CLAM and RS-8-samples). From left to right, the slide, attention heatmap generated by the CLAM, and attention heatmap produced by the RS-8-samples.

## 3. RESULTS

Table 1 and Figure 4 present the random sampling method results for CAMELYON16 and TUPAC16. The model achieves the best performance on CAMELYON16 with 30% of samples. In contrast, the peak performance on TUPAC16 occurs with only eight samples.

Table 1. The 10-fold average performance (±std) of different methods trained on the two datasets in terms of AUC.

| Method | AUC (CAMELYON16) | AUC (TUPAC16) |
|---|---|---|
| **RS-8-samples** | 0.7353 ± 0.0299 | **0.7892 ± 0.013** |
| **RS-2%** | 0.8611 ± 0.0420 | 0.7775 ± 0.012 |
| **RS-6%** | 0.8992 ± 0.0335 | 0.7741 ± 0.020 |
| **RS-10%** | 0.9135 ± 0.0313 | 0.7698 ± 0.018 |
| **RS-30%** | **0.9216 ± 0.0278** | 0.7679 ± 0.022 |
| **RS-60%** | 0.9199 ± 0.0196 | 0.7654 ± 0.020 |
| **RS-90%** | 0.8993 ± 0.0390 | 0.7639 ± 0.024 |
| **CLAM (RS-100%)** | 0.9054 ± 0.0238 | 0.7605 ± 0.022 |

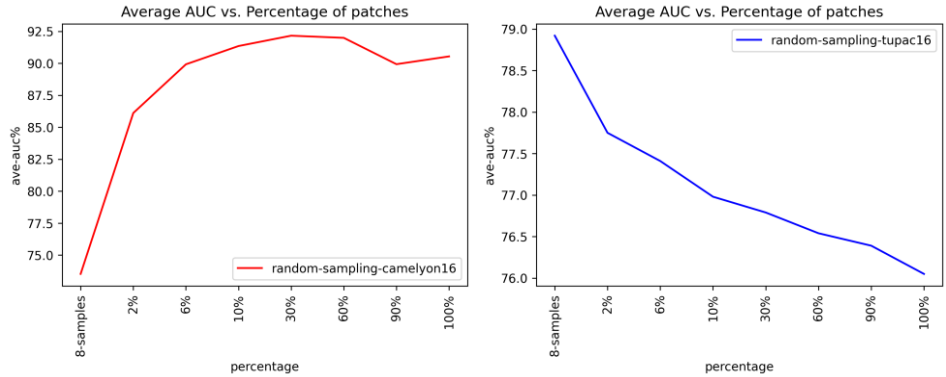

Figure 4. The 10-fold average performance of the model trained on the two datasets in terms of AUC for variable percentages of patches.

We display attention heatmaps to evaluate the interpretability of the methods for both normal and tumor slides. Upon observing Figure 2, it is evident that for CAMELYON16 dataset, the RS-30% network focuses more on tumor tissues to classify tumor slides. In contrast, when analyzing normal slides, the CLAM model concentrates heavily on fat regions, whereas the RS-30% network primarily focuses on normal tissue. This shows improvement on the interpretability of the model. Conversely, for TUPAC16 dataset as seen in Figure 3, the heatmaps generated by both models for different classes are very similar. The results of E2E are presented in Table 2 for both datasets. We could notice that there are improvements in both datasets even though they are not considerable. To provide a complete analysis of the results, we rendered AUC curves for the methods in Figure 5. It is worth mentioning that we did not provide heatmaps generated by the models, as there were no significant differences between them.

## 4. DISCUSSION

In this paper, we thoroughly investigated random sampling of whole-slide image patches for the training of multiple instance learning method. Our results show that on CAMELYON16 dataset, where tumor tissues are only present in a small fraction of each tumor slide, optimal model performance is achieved by sampling 30% of patches during training. On the other hand, for TUPAC16, where image features associated with tumor proliferation scores are more uniformly distributed throughout the slides, sampling only 8 patches at training time is sufficient for the network to achieve optimal performance. Moreover, these results emphasize the importance of carefully selecting the number of instances for each dataset to achieve high performance with the assistance of random sampling, as noted in [4, 10]. However, we observed that random sampling enhanced the model's interpretability on CAMELYON16, as it prevents overfitting on this small dataset. Second, our study found that end-to-end MIL produced better results than using pre-extracted features.

## 5. CONCLUSIONS

In summary, this paper explores the impact of random sampling on MIL for Whole Slide Images classification. Optimal performance improvements vary for different datasets, with a 1.7% enhancement using 30% of patches on CAMELYON16 and a 3.7% improvement with only eight samples on TUPAC16. In addition, model interpretability effects are dataset-dependent, with interpretability being enhanced on CAMELYON16 but remaining unaffected on TUPAC16. Finally, end-to-end training with 1024 samples improved the performance (1.7% on CAMELYON16 and 1.1% on TUPAC16). Overall, the research provides insights into optimizing MIL for computational pathology.

## 6. ACKNOWLEDGEMENTS

Table 2. The performance of different methods trained, using pre-extracted features and in the end-to-end fashion in terms of AUC. The 95% confidence intervals based on 2000 bootstrapping results are shown in parentheses.

| Method | AUC (CI) |
|---|---|
| CLAM-CAMELYON | 0.910 (0.857 - 0.958) |
| E2E-CLAM-CAMELYON | **0.926 (0.873 - 0.969)** |
| CLAM-TUPAC | 0.783 (0.729 - 0.831) |
| E2E-CLAM-TUPAC | **0.792 (0.738 - 0.841)** |

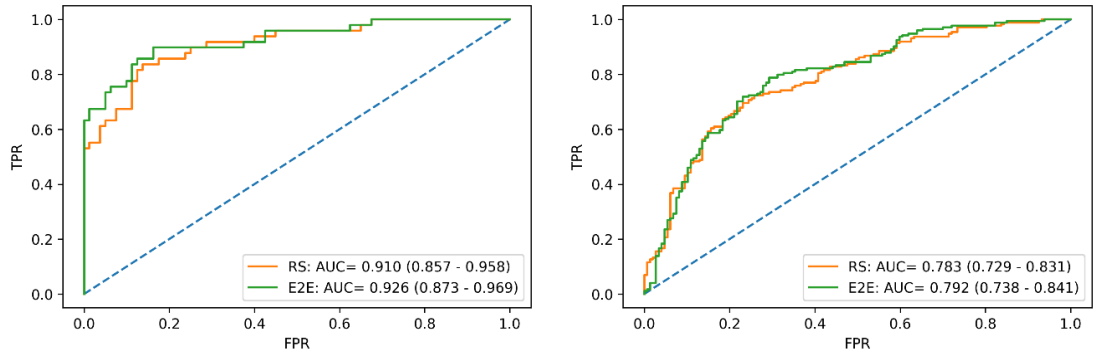

Figure 5. AUC curve for different methods. Left: CLAM-CAMELYON, and Right: CLAM-TUPAC.